%% file: iclr2025_conference.tex
\documentclass{article} 
\usepackage{iclr2025_conference,times}

\input{math_commands.tex}

\usepackage{hyperref}
\usepackage{url}
\usepackage{enumitem} 

\usepackage{graphicx}
\usepackage{subcaption} 
\usepackage{booktabs} 
\newcommand{\name}{SAGE-KV\xspace} 
\usepackage{xspace}
\usepackage{wrapfig}

\newcommand{\xhdr}[1]{{\noindent\bfseries #1}.}
\newcommand{\xhdrnd}[1]{{\noindent\bfseries #1}} 

\usepackage{threeparttable} 

\title{LLMs Know What to Drop: Self-Attention Guided KV Cache Eviction for Efficient Long-Context Inference}

\author{Guangtao Wang~\thanks{Corresponding author.  xjtuwgt@gmail.com, guangtao.wang@sambanovasystems.com}, Shubhangi Upasani, Chen Wu, Darshan Gandhi, Jonathan Li, \\ \textbf{Changran Hu, Bo Li, Urmish Thakker} \\
SambaNova Systems, Inc\\
Palo Alto, CA 94303, USA \\
\texttt{\url{https://sambanova.ai/}} 
}

\iclrfinalcopy 
\begin{document}

\maketitle

\begin{abstract}

Efficient long-context inference is critical as large language models (LLMs) adopt context windows of ranging from 128K to 1M tokens. However, the growing key-value (KV) cache and the high computational complexity of attention create significant bottlenecks in memory usage and latency. In this paper, we find that attention in diverse long-context tasks exhibits sparsity, and LLMs implicitly ``know" which tokens can be dropped or evicted at the head level after the pre-filling stage. Based on this insight, we propose Self-Attention Guided Eviction~(\name), a simple and effective KV eviction cache method for long-context inference. After prefilling, our method performs a one-time top-$k$ selection at both the token and head levels to compress the KV cache, enabling efficient inference with the reduced cache. Evaluations on LongBench and three long-context LLMs (Llama3.1-8B-Instruct-128k, Llama3-8B-Prolong-512k-Instruct, and Qwen2.5-7B-Instruct-128k) show that \name maintains accuracy comparable to full attention while significantly improving efficiency. Specifically, \name achieves 4x higher memory efficiency with improved accuracy over the static KV cache selection method StreamLLM, and 2x higher memory efficiency with better accuracy than the dynamic KV cache selection method Quest.

\end{abstract}

\section{Introduction}\label{sec: intro}

Long-context Large Language Models (LLMs) are essential for tasks like summarization, multi-hop reasoning, question answering, code understanding, personalized chatbots, recommendations, and in-context learning~\citep{zhou2024survey,li2024survey,gpt4o,team2024gemini}. However, their deployment is limited by high computational costs, driven by the KV cache’s memory demands and attention computation latency~\citep{li2024survey}. As attention latency grows with KV cache size, efficient memory and computation management are crucial for real-world feasibility. Addressing these challenges is key to fully leveraging long-context LLMs.

Recent efforts to reduce KV cache requirements and accelerate inference in long-context LLMs have gained increasing attention, mainly by exploiting attention sparsity~\citep{zhang2023h2o,liu2024retrievalattention}. Sparsity patterns fall into two main categories: \textbf{static}~\citep{xiaoefficient,xiao2024duoattention} and \textbf{dynamic}~\citep{xiao2024infllm,tangquest,liu2024retrievalattention,sun2024shadowkv}. Static sparsity methods predefine token selection rules, avoiding runtime computation and enabling faster inference. For instance, StreamLLM~\citep{xiaoefficient} retains sink tokens (early context) and recent tokens, reducing KV cache size without additional selection overhead. 

Dynamic sparsity methods adaptively select representative tokens per generation step, often yielding higher accuracy but at increased computational cost. They require careful hyperparameter tuning (e.g., chunk size in InfLLM~\citep{xiao2024infllm}, or ANN index construction in RetrievalAttention~\citep{liu2024retrievalattention}) and must retain the full KV cache as a candidate pool, limiting memory savings. Although offloading KV caches to the CPU reduces GPU memory usage, it incurs high retrieval latency~\citep{sun2024shadowkv}. As a result, dynamic methods demand sophisticated CPU-GPU coordination~\citep{xiao2024infllm,lee2024infinigen,he2024fastdecode,liu2024retrievalattention}, increasing implementation complexity.

\begin{figure}[!h]
    \vspace{-18pt}
    \centering
    \includegraphics[width=0.6\linewidth]{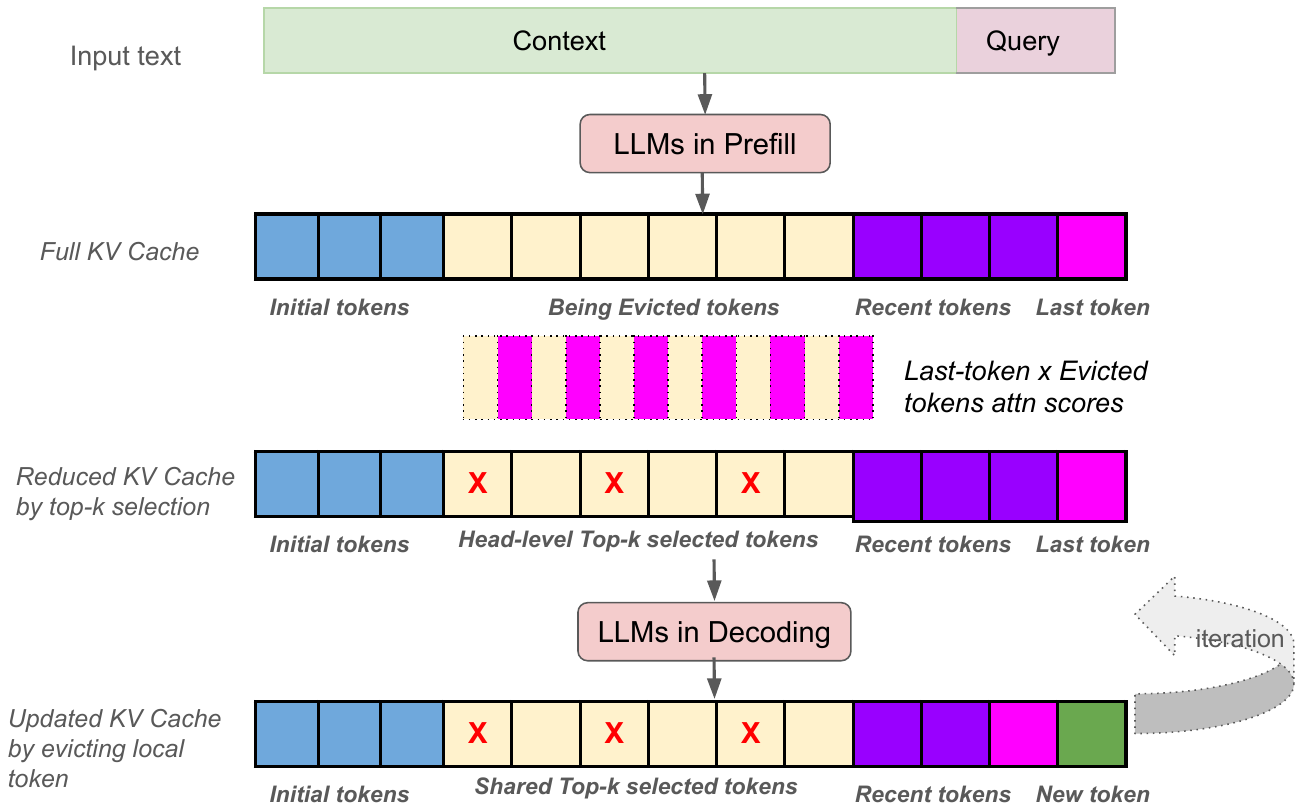} 
    \caption{\textbf{The illustration of \name with the single-pass KV cache selection}. The full KV cache consists of four parts: initial tokens, tokens for eviction, recent tokens, and the last token. To construct a reduced KV cache, we select the top-$k$ evicted tokens based on their attention scores with the last token and concatenate them with the initial and recent tokens. The updated cache is used for continuous token generation, with each new token (green) added to the recent tokens, updating the recent window for the next step.}
    \label{fig:framework}
    \vspace{-15pt}
\end{figure}

In this paper, we tackle efficient long-context inference in LLMs by leveraging the observation that, after the pre-filling stage, LLMs naturally focus on critical information. We analyze the sparsity of the attention score and find that the attention heads selectively highlight important tokens. This insight motivates our optimization of KV cache compression at the head level. Based on this, we propose \textbf{\underline{S}elf-\underline{A}ttention \underline{G}uided \underline{E}viction for KV Cache} (\name) (Fig.~\ref{fig:framework}), a novel method that uses attention scores to guide eviction of KV cache, significantly improving inference efficiency while preserving precision.

\name employs a single-pass token-level KV cache selection strategy, compressing the KV cache once after the pre-filling stage using attention scores (Fig.~\ref{fig:framework}). Only the compressed KV cache is retained, eliminating redundant KV selection during token generation, unlike block-level dynamic methods. This reduces computational overhead while preserving essential information for inference. By retaining only the most relevant tokens, \name ensures both efficiency and accuracy. It combines the fixed sparsity of static methods—benefiting from their single-pass processing and structured cache—with the context-adaptive selection of dynamic methods. This synergy enables \name to achieve superior efficiency while maintaining or even improving performance over existing dynamic block-level KV selection approaches. 

Extensive experiments across long-context LLMs and benchmarks validate these advantages. \name achieves nearly 4x higher memory efficiency and improved accuracy compared to the static KV eviction method StreamLLM~\cite{xiaoefficient}, and 2x higher memory efficiency over the dynamic block-level KV eviction method Quest~\cite{tangquest}. This simple yet effective approach accelerates long-context inference while remaining easy to integrate, offering a promising solution to the challenges of long-context LLM inference.

\section{Self-Attention Guided KV Cache Eviction}\label{sec:method}

Given a long input sequence $s = [t_{i}]_{i=1}^{N}$ of length $N$, consisting of a context followed by a query, the pre-filling step produces the full key-value cache for layer $l$: $\mathbf{P}^{l} = [(\mathbf{k}_i^{l}, \mathbf{v}_i^{l})]_{i=1}^{N}$. As illustrated in Fig.~\ref{fig:framework}, the proposed method proceeds as follows.

\xhdr{Step 1: Full KV Cache Partition}  We divide the KV cache $\mathbf{P}^{l}$ into four parts: (1) initial tokens or sink tokens $\mathbf{S}^{l} = \mathbf{P}^{l}_{1:S} $ with length $S$, (2) evicted tokens $\mathbf{E}^{l} = \mathbf{P}^{l}_{S+1:S+E}$ with length $E$, (3) recent tokens $\mathbf{R} = \mathbf{P}^{l}_{S+E+1:N-1}$ with length $R$ = $N - 1 - (S+E)$ and (4) the last token’s KV cache $\mathbf{P}^{l}_{N}$. Sink and recent tokens are retained separately, as attention analysis shows that initial and most recent tokens typically receive higher attention scores across all heads~\citep{xiaoefficient}, which refers to this as the ``attention sink".

\xhdr{Step 2: Representative Token/KV Cache Selection} We select representative KV cache entries based on the attention scores between the last token of the input sequence and the evicted tokens.  Let $\mathbf{q}^{l} \in \mathcal{R}^{H_q\times d_h}$ denote the query vector corresponding to the last token’s KV cache $\mathbf{P}^{l}_{N}$, where $H_q$ is the number of query heads, $d_h$ is the head dimension, and $H_q \times d_h = d$, the hidden dimension. In decoder-only LLMs, the last token’s hidden representation often serves as an embedding for the entire input sequence~\citep{lee2024nv,behnamghader2024llm2vec}. Thus, $\mathbf{q}^{l}$ acts as a representative embedding for the full sequence.

For each layer $l$, we use $\mathbf{q}^{l}$ to select the top-$k$ KV cache entries, $\mathbf{E}_{\text{top}_k}^{l}$, based on the attention scores with $\mathbf{E}^{l}$. This yields $H_q$ groups of the top-$k$ KV caches, forming a representative set of key-value pairs for the next-token generation.

\xhdr{Step 3: Reduced KV Cache Construction} The reduced KV cache $\mathbf{C}$ is formed by concatenating sink token KV cache, selected top-$k$ token KV cache, recent token KV cache as well as the last KV cache, resulting in $\mathbf{C} = \text{Concat}(\mathbf{S}, \mathbf{E}_{top_k}, \mathbf{R}, \mathbf{P}^{l}_{N})$ with total length $S + k + R + 1$.

\xhdr{Step 4: Generation/Decoding} The output is generated using the reduced KV cache $\mathbf{C}$. Each new token’s KV pair is added to the recent window $\mathbf{R}$, evicting the oldest entry in $\mathbf{R}$ to maintain its size. This process repeats until generation completes.

\section{Experiments and Results}
\label{sec:experiment}

\subsection{Experimental Setup}

\xhdr{Benchmarks and long context LLMs} We assess long-context processing with LongBench~\citep{bai2023longbench} on tasks like QA, summarization, retrieval, and code analysis. Experiments span Llama3.1-8B-Instruct (128k)\citep{dubey2024llama}, Llama-3-8B-ProLong-512k-Instruct\citep{gao2024train}, and Qwen2.5-7B-Instruct (128k)~\citep{hui2024qwen2}, ensuring broad model evaluation.

\xhdr{Baselines} We compare our method with the following: (1) \textbf{Full-Attention Models}: which use standard full attention; (2) \textbf{Hugging Face StreamLLM}~\citep{xiaoefficient}: the official implementation; (3) \textbf{Our StreamLLM Implementation}: which addresses position conflicts\footnote{\url{https://github.com/huggingface/transformers/issues/35350}} in the Hugging Face version by introducing (a) $\text{StreamLLM}_R$, which stores pre-RoPE KV cache and applies local window relative positional encoding, and (b) $\text{StreamLLM}_{\text{Abs}}$, which uses absolute positional encoding; (See Appendix) (4) \textbf{Quest}~\citep{tangquest}: a block-wise top-$k$ selection method; and (5) \textbf{InfLLM}~\citep{xiao2024infllm}: which integrates sink tokens, recent tokens, and block-wise top-$k$ selection. See Appendix for \name implementation details.

\begin{table}[h!]
\centering
\caption{Accuracy Comparison of KV Cache Eviction Methods on LongBench}
\label{tab:long_bench}
\resizebox{0.95\textwidth}{!}{%
\begin{tabular}{@{}lcccccccccccccccc@{}}
\toprule
Method & 2wikimqa & gov.rep & lcc & mulfqa & mul.news & nar.qa & pass.ret & qasper & rep-p & trec & hotpotqa & musique & qmsum & samsum & tri.qa & Average \\ \midrule
\multicolumn{17}{c}{Llama3.1-8B-Instruct (128k)} \\
\midrule
FullAttention  &  50.01 & 34.66 & 65.52 & 56.85 & 27.11 & 31.38 & 100.0 & 46.6 & 58.3 & 73.0 & 58.1 & 32.52 & 25.19 & 43.75 & 92.11 & 53.01 \\
\midrule
StreamLLM$_{HF}$  & 49.47 & 32.66 & 65.37 & 55.54 & 27.05 & 30.98 & 79.0 & 46.27 & 54.87 & 65.0 & 56.20 & 30.35 & 24.45 & 42.89 & 91.93 & 50.14 \\
StreamLLM$_{R}$ & 49.08 & 33.68 & 65.48 & 55.30 & 27.14 & 31.14 & 83.0 & 45.72 & 55.38 & 73.0 & 57.44 & 30.37 & 24.61 & 43.69 & 91.97 & 51.13  \\
StreamLLM$_{Abs}$ & 48.58 & 33.96 & 65.33 & 55.06 & 27.14 & 30.6 & 84.5 & 45.71 & 56.11 & 73.0 & 57.58 & 30.08 & 24.37 & 43.65 & 92.04 & 51.18 \\
Quest & 45.83 & 35.25 & 64.8 & 55.36 & 27.25 & 28.14 & 99.5 & 44.95 & 59.69 & 70.5 & 55.17 & 31.69 & 25.40 & 42.75 & 82.83 & 51.27 \\
InfLLM & 45.61 & 34.42 & 67.2 & 51.21 & 27.69 & 23.82 & 92.0 & 44.75 & 64.44 & 69.0 & 50.65 & 23.74 & 24.24 & 43.4 & 92.16 & 50.29  \\
\name (Ours) &  48.33 & 33.98 & 65.01 & 56.16 & 26.95 & 31.42 & 100.0 & 46.21 & 55.46 & 73.0 & 58.08 & 33.29 & 24.32 & 43.56 & 91.64 & \textbf{\underline{52.49}}   \\
\midrule
\multicolumn{17}{c}{Llama-3-8B-ProLong-512k-Instruct}\\
\midrule
FullAttention  & 24.71 & 33.23 & 66.01 & 53.91 & 27.67 & 29.66 & 100.0 & 31.05 & 67.4 & 73.5 & 36.24 & 13.96 & 25.53 & 43.19 & 90.01 & 47.74 \\
\midrule
StreamLLM$_{HF}$  & 24.06 & 30.78 & 66.11 & 53.16 & 27.73 & 27.79 & 67.5 & 30.66 & 66.0 & 74.0 & 34.64 & 12.57 & 23.88 & 43.55 & 90.48 & 44.86 \\
StreamLLM$_{R}$  & 25.26 & 32.38 & 65.95 & 51.62 & 27.58 & 26.36 & 68.0 & 30.65 & 66.49 & 74.0 & 35.31 & 11.84 & 24.66 & 43.53 & 89.51 & 44.88 \\
StreamLLM$_{Abs}$  & 25.63 & 32.48 & 65.95 & 52.74 & 27.54 & 26.67 & 68.5 & 30.54 & 65.8 & 74.0 & 34.28 & 11.52 & 24.60 & 43.12 & 89.38 & 44.85 \\
Quest & 24.86 & 31.90 & 65.20 & 50.75 & 28.13 & 26.18 & 99.5 & 29.52 & 67.33 & 72.5 & 35.58 & 14.89 & 25.77 & 43.11 & 86.76 & 46.80 \\
InfLLM & 24.84 & 28.37 & 38.10 & 68.31 & 25.43 & 49.18 & 14.34 & 22.25 & 66.4 & 49.5 & 29.56 & 23.16 & 41.17 & 75.50 & 90.06 & 43.08 \\
\name (Ours)  & 25.57 & 32.08 & 65.96 & 54.74 & 27.72 & 29.90 & 100.0 & 30.03 & 67.39 & 73.0 & 36.48 & 13.76 & 25.0 & 42.94 & 89.98 & \textbf{\underline{47.64}} \\
\midrule
\multicolumn{17}{c}{Qwen2.5-7B-Instruct (128k)}\\
\midrule
FullAttention    & 46.11 & 33.46 & 62.03 & 51.77 & 25.91 & 28.79 & 100.0 & 44.79 & 66.83 & 73.5 & 58.38 & 29.23 & 24.32 & 47.92 & 88.93 & 52.13      \\
\midrule
StreamLLM$_{HF}$ & 11.86 & 29.70 & 61.92 & 22.64 & 23.08 & 3.37 & 59.58 & 10.09 & 64.87 & 72.5 & 11.33 & 7.05 & 16.88 & 46.45 & 88.19 & 35.3 \\
StreamLLM$_{R}$ & 44.92 & 33.2 & 62.18 & 49.10 & 25.73 & 25.85 & 70.0 & 44.79 & 65.88 & 72.0 & 50.73 & 24.83 & 23.20 & 47.73 & 89.41 & 48.64 \\
StreamLLM$_{Abs}$ & 45.46 & 33.23 & 61.82 & 49.63 & 25.71 & 23.65 & 71.5 & 44.79 & 65.29 & 72.5 & 52.61 & 24.9 & 23.39 & 47.45 & 88.64 & 48.70 \\
Quest & 45.70 & 32.73 & 60.45 & 50.41 & 25.77 & 26.34 & 97.83 & 44.70 & 59.95 & 73.0 & 57.65 & 28.79 & 24.50 & 47.02 & 86.78 & 50.77 \\
InfLLM & 43.12 & 32.90 & 52.51 & 60.45 & 24.65 & 48.23 & 28.11 & 22.76 & 63.63 & 69.0 & 43.27 & 23.05 & 45.20 & 66.5 & 88.45 & 47.46 \\
\name (Ours) & 45.81 & 32.53 & 61.72 & 50.29 & 25.79 & 28.05 & 100.0 & 44.21 & 63.0 & 72.0 & 55.90 & 29.16 & 23.17 & 47.51 & 88.65 & \textbf{\underline{51.19}} \\
 \bottomrule
\end{tabular}%
}
\end{table}

\subsection{Results}

\xhdrnd{Accuracy comparison} on LongBench is represented in in Table~\ref{tab:long_bench}, which shows that:
\begin{itemize}[leftmargin=2pt, parsep=0pt, topsep=0pt]
    \item \name achieves accuracy comparable to full attention by applying self-attention-guided KV cache selection after pre-filling. Unlike per-token generation based selection, it selects KV entries in a single pass using the last input token’s attention scores, leveraging LLMs’ inherent ability to prioritize key tokens for effective answer generation. 
    
    \item StreamLLM’s static sparse KV selection degrades accuracy in long-context LLMs by discarding essential information from the middle of the input. Our implementation outperforms the Hugging Face (HF) version, with significantly better results on Qwen2.5-7B-Instruction and slightly higher accuracy on Llama3.1-8B-Instruct. The improvement likely stems from a flaw in HF’s RoPE rotation reported in~\cite{hf_github_bug}, which misaligns key cache positions and increases relative token distances, particularly affecting Qwen2.5. Our implementation corrects this, ensuring more stable performance and underscoring the importance of precise implementation for StreamLLM, especially in position-sensitive models.

    \item Dynamic sparse KV selection methods like InfLLM and Quest retrieve relevant blocks per step but underperform compared to \name. Their block-wise top-$k$ selection, relying on pooled vectors over blocks, weakens critical token retention~\citep{liu2024retrievalattention,tangquest}, leading to lower accuracy~\citep{gao2024train}. While block-level selection reduces latency, it struggles to preserve essential tokens. In contrast, \name’s token-level selection better approximates full attention, achieving higher accuracy and highlighting the importance of precise token selection for efficient KV cache management.
\end{itemize}

\begin{wrapfigure}{R}{0.35\textwidth}  
    \centering
    \vspace{-18pt}
    \includegraphics[width=0.36\textwidth]{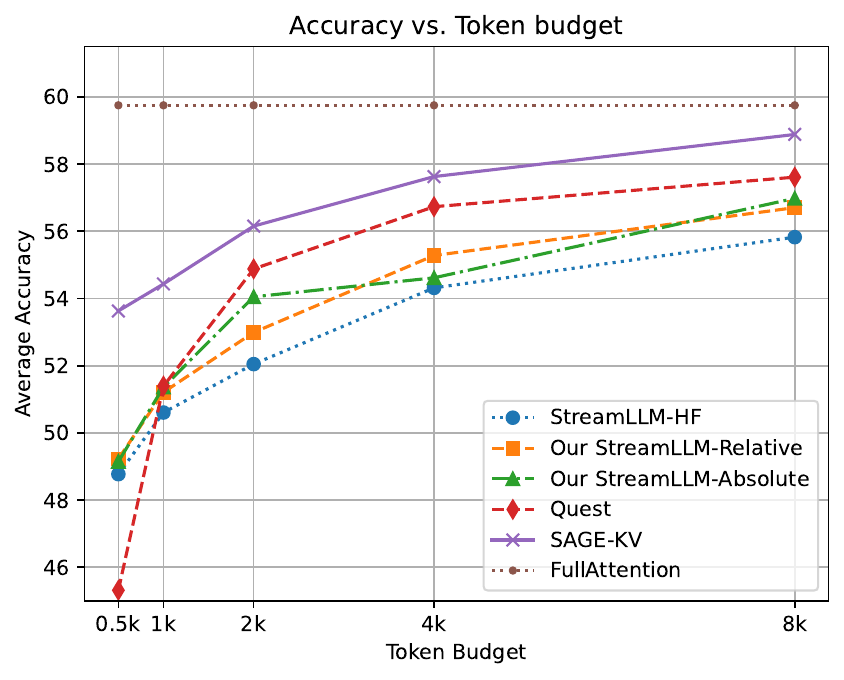}
    \caption{Token Budget Analysis.}
    \label{fig:budget}
    \vspace{-18pt}
\end{wrapfigure}

\xhdr{Memory efficiency analysis} We evaluate KV cache eviction methods on Llama3.1-8B-Instruct, measuring average accuracy across eight LongBench tasks (See Appendix) under token budgets  B  of 0.5k, 1k, 2k, 4k, and 8k. Results in Fig.~\ref{fig:budget} reveal the following insights.


(1) Performance improves across all methods as the token budget increases, indicating better information retention and higher accuracy.

(2) \name consistently outperforms StreamLLM’s static sparse KV selection across all token budgets by reducing the significant information loss caused by discarding middle sections. Instead, \name employs a self-attention-guided top-$k$ token selection strategy, preserving critical information within the same budget. This demonstrates the effectiveness of our adaptive KV eviction method.
    
(3) With a 2k token budget, \name achieves the same accuracy as StreamLLM at 8k, improving memory efficiency by $\sim$4x on LongBench tasks. This highlights the advantage of attention-guided KV eviction in balancing performance and memory usage.
    
(4) \name achieves $\sim$2x memory efficiency by matching Quest’s performance with half the token budget (4k vs. 8k). Unlike Quest’s coarse chunk-level top-$k$ selection, \name employs token-level selection for more precise context retention, leveraging LLMs’ ability to prioritize important tokens.

\section{Conclusion and Future Work}

We introduce \name, an efficient KV cache eviction method for long-context LLM inference. Exploiting attention sparsity, \name compresses the KV cache after prefilling for direct use in generation. Our experiments show that \name matches the inference speed of static sparse methods while preserving accuracy close to full attention. It achieves \textit{$\sim$4x higher memory efficiency} and \textit{greater accuracy} than the static eviction method StreamLLM, and \textit{$\sim$2x higher memory efficiency} with \textit{better accuracy} than the dynamic method Quest. Additionally, \name seamlessly integrates with popular LLM frameworks, including Hugging Face Transformers, Meta’s LLaMA, and Alibaba’s Qwen, ensuring broad applicability.

\xhdr{Future Work} Current long-context tasks mainly involve short outputs, such as question answering and retrieval. However, for long-text generation, a single top-k selection may be insufficient. Future work will evaluate our method on long-output benchmarks and introduce interval-based updates, where the LLM periodically refreshes selected key tokens to improve coherence and relevance.



\clearpage
\newpage
\appendix
\section{Appendix}

\subsection{StreamLLM with Relative and Absolute Positions}\label{append:stream_position}

\begin{figure}[!h] 
    \centering
    \includegraphics[width=0.8\textwidth]{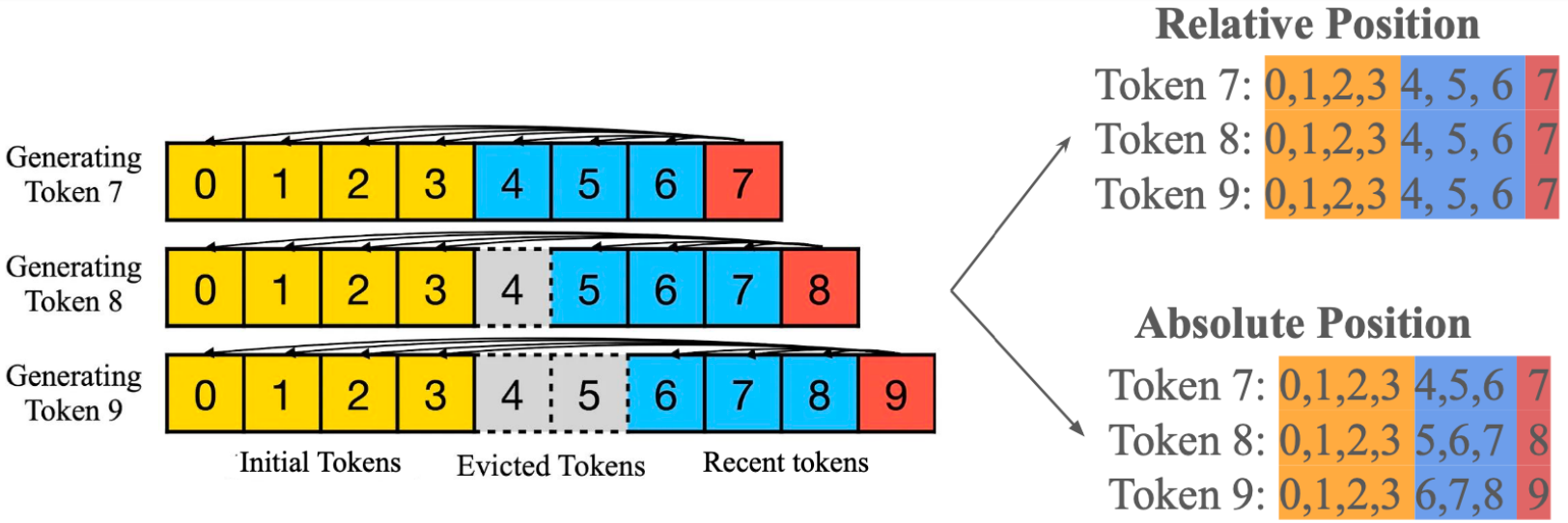} 
    \caption{Implementation of Absolute and Relative Positioning in StreamLLM. Relative position assigns indices within StreamLLM’s sliding window, dynamically shifting as the window moves to maintain a bounded range. In contrast, absolute position assigns a fixed index to each token based on its original sequence, increasing continuously as new tokens are added.}
    \label{fig:streamLLM_position} 
\end{figure}

\subsection{Implementation Details of \name with LLMs}

\xhdr{Hyper-parameter settings} For our method \name, suppose that the token budget is $B$ and the query group number is $G = h_{q} / h_{v}$ where $h_{q}$ and $h_{kv}$ are the query head number and ke/value head number, respectively, we set the sink window as $B/4$, and $k$ = $\frac{B}{2G}$, and the recent window size as $B/4$. For Llama3.1-8B-Instruct and Llama-3-8B-ProLong-512k-Instruct, we set $B$ = 8192, and thus sink window size = 2048 and $k$ = 1024, the local window size = 2048. For Qwen2.5, since the query group number is 7, we set $B$ = 8192, thus sink window size = 2048, $k$ = 512 and the local window size = 8192 - 2048 - 7 $\cdot$ 512 = 2560. For other baselines, we set the same token budget $B$ 8192. Absolute positioning is applied in \name to the reduced KV cache to maintain token order.

\xhdr{Task names for token budget analysis} We use the following eight LongBench tasks for hyper-parameter analysis, as in~\citep{tangquest}: ``gov-report", ``multifieldqa-en", ``narrativeqa", ``passage-retrieval-en", ``qasper", ``repobench-p", ``hotpotqa", and ``triviaqa".

\end{document}

%% file: math_commands.tex
\usepackage{amsmath,amsfonts,bm}

\def\eqref#1{equation~\ref{#1}}

\def\1{\bm{1}}

\DeclareMathAlphabet{\mathsfit}{\encodingdefault}{\sfdefault}{m}{sl}
\SetMathAlphabet{\mathsfit}{bold}{\encodingdefault}{\sfdefault}{bx}{n}













%% file: iclr2025_conference.bbl
\begin{thebibliography}{20}
\providecommand{\natexlab}[1]{#1}
\providecommand{\url}[1]{\texttt{#1}}
\expandafter\ifx\csname urlstyle\endcsname\relax
  \providecommand{\doi}[1]{doi: #1}\else
  \providecommand{\doi}{doi: \begingroup \urlstyle{rm}\Url}\fi

\bibitem[Bai et~al.(2023)Bai, Lv, Zhang, Lyu, Tang, Huang, Du, Liu, Zeng, Hou, et~al.]{bai2023longbench}
Yushi Bai, Xin Lv, Jiajie Zhang, Hongchang Lyu, Jiankai Tang, Zhidian Huang, Zhengxiao Du, Xiao Liu, Aohan Zeng, Lei Hou, et~al.
\newblock Longbench: A bilingual, multitask benchmark for long context understanding.
\newblock \emph{arXiv preprint arXiv:2308.14508}, 2023.

\bibitem[BehnamGhader et~al.(2024)BehnamGhader, Adlakha, Mosbach, Bahdanau, Chapados, and Reddy]{behnamghader2024llm2vec}
Parishad BehnamGhader, Vaibhav Adlakha, Marius Mosbach, Dzmitry Bahdanau, Nicolas Chapados, and Siva Reddy.
\newblock Llm2vec: Large language models are secretly powerful text encoders.
\newblock \emph{arXiv preprint arXiv:2404.05961}, 2024.

\bibitem[Dubey et~al.(2024)Dubey, Jauhri, Pandey, Kadian, Al-Dahle, Letman, Mathur, Schelten, Yang, Fan, et~al.]{dubey2024llama}
Abhimanyu Dubey, Abhinav Jauhri, Abhinav Pandey, Abhishek Kadian, Ahmad Al-Dahle, Aiesha Letman, Akhil Mathur, Alan Schelten, Amy Yang, Angela Fan, et~al.
\newblock The llama 3 herd of models.
\newblock \emph{arXiv preprint arXiv:2407.21783}, 2024.

\bibitem[Gao et~al.(2024)Gao, Wettig, Yen, and Chen]{gao2024train}
Tianyu Gao, Alexander Wettig, Howard Yen, and Danqi Chen.
\newblock How to train long-context language models (effectively).
\newblock \emph{arXiv preprint arXiv:2410.02660}, 2024.

\bibitem[He \& Zhai(2024)He and Zhai]{he2024fastdecode}
Jiaao He and Jidong Zhai.
\newblock Fastdecode: High-throughput gpu-efficient llm serving using heterogeneous pipelines.
\newblock \emph{arXiv preprint arXiv:2403.11421}, 2024.

\bibitem[{Hugging Face Issue}(2024)]{hf_github_bug}
{Hugging Face Issue}.
\newblock {SinkCache (StreamLLM) implemented over Post-RoPE Key cache might result in confused position for inference}, 2024.
\newblock URL \url{https://github.com/huggingface/transformers/issues/35350}.

\bibitem[Hui et~al.(2024)Hui, Yang, Cui, Yang, Liu, Zhang, Liu, Zhang, Yu, Lu, et~al.]{hui2024qwen2}
Binyuan Hui, Jian Yang, Zeyu Cui, Jiaxi Yang, Dayiheng Liu, Lei Zhang, Tianyu Liu, Jiajun Zhang, Bowen Yu, Keming Lu, et~al.
\newblock Qwen2. 5-coder technical report.
\newblock \emph{arXiv preprint arXiv:2409.12186}, 2024.

\bibitem[Josh~Achiam(2023)]{gpt4o}
Sandhini Agarwal Lama Ahmad Ilge Akkaya Florencia Leoni Aleman Diogo Almeida Janko Altenschmidt Sam Altman Shyamal Anadkat et~al. Josh~Achiam, Steven~Adler.
\newblock Gpt-4 technical report.
\newblock \emph{arXiv preprint arXiv:2303.08774}, 2023.

\bibitem[Lee et~al.(2024{\natexlab{a}})Lee, Roy, Xu, Raiman, Shoeybi, Catanzaro, and Ping]{lee2024nv}
Chankyu Lee, Rajarshi Roy, Mengyao Xu, Jonathan Raiman, Mohammad Shoeybi, Bryan Catanzaro, and Wei Ping.
\newblock Nv-embed: Improved techniques for training llms as generalist embedding models.
\newblock \emph{arXiv preprint arXiv:2405.17428}, 2024{\natexlab{a}}.

\bibitem[Lee et~al.(2024{\natexlab{b}})Lee, Lee, Seo, and Sim]{lee2024infinigen}
Wonbeom Lee, Jungi Lee, Junghwan Seo, and Jaewoong Sim.
\newblock $\{$InfiniGen$\}$: Efficient generative inference of large language models with dynamic $\{$KV$\}$ cache management.
\newblock In \emph{18th USENIX Symposium on Operating Systems Design and Implementation (OSDI 24)}, pp.\  155--172, 2024{\natexlab{b}}.

\bibitem[Li et~al.(2024)Li, Li, Tian, Tang, Xu, Chen, Hu, Dong, Li, and Chen]{li2024survey}
Haoyang Li, Yiming Li, Anxin Tian, Tianhao Tang, Zhanchao Xu, Xuejia Chen, Nicole Hu, Wei Dong, Qing Li, and Lei Chen.
\newblock A survey on large language model acceleration based on kv cache management.
\newblock \emph{arXiv preprint arXiv:2412.19442}, 2024.

\bibitem[Liu et~al.(2024)Liu, Chen, Lu, Jiang, Han, Zhang, Chen, Zhang, Ding, Zhang, et~al.]{liu2024retrievalattention}
Di~Liu, Meng Chen, Baotong Lu, Huiqiang Jiang, Zhenhua Han, Qianxi Zhang, Qi~Chen, Chengruidong Zhang, Bailu Ding, Kai Zhang, et~al.
\newblock Retrievalattention: Accelerating long-context llm inference via vector retrieval.
\newblock \emph{arXiv preprint arXiv:2409.10516}, 2024.

\bibitem[Sun et~al.(2024)Sun, Chang, Bao, Zheng, Zheng, Liu, Dong, Chi, and Chen]{sun2024shadowkv}
Hanshi Sun, Li-Wen Chang, Wenlei Bao, Size Zheng, Ningxin Zheng, Xin Liu, Harry Dong, Yuejie Chi, and Beidi Chen.
\newblock Shadowkv: Kv cache in shadows for high-throughput long-context llm inference.
\newblock \emph{arXiv preprint arXiv:2410.21465}, 2024.

\bibitem[Tang et~al.(2024)Tang, Zhao, Zhu, Xiao, Kasikci, and Han]{tangquest}
Jiaming Tang, Yilong Zhao, Kan Zhu, Guangxuan Xiao, Baris Kasikci, and Song Han.
\newblock Quest: Query-aware sparsity for efficient long-context llm inference.
\newblock In \emph{Forty-first International Conference on Machine Learning}, 2024.

\bibitem[Team et~al.(2024)Team, Georgiev, Lei, Burnell, Bai, Gulati, Tanzer, Vincent, Pan, Wang, et~al.]{team2024gemini}
Gemini Team, Petko Georgiev, Ving~Ian Lei, Ryan Burnell, Libin Bai, Anmol Gulati, Garrett Tanzer, Damien Vincent, Zhufeng Pan, Shibo Wang, et~al.
\newblock Gemini 1.5: Unlocking multimodal understanding across millions of tokens of context.
\newblock \emph{arXiv preprint arXiv:2403.05530}, 2024.

\bibitem[Xiao et~al.(2024{\natexlab{a}})Xiao, Zhang, Han, Xiao, Lin, Zhang, Liu, and Sun]{xiao2024infllm}
Chaojun Xiao, Pengle Zhang, Xu~Han, Guangxuan Xiao, Yankai Lin, Zhengyan Zhang, Zhiyuan Liu, and Maosong Sun.
\newblock Infllm: Training-free long-context extrapolation for llms with an efficient context memory.
\newblock In \emph{The Thirty-eighth Annual Conference on Neural Information Processing Systems}, 2024{\natexlab{a}}.

\bibitem[Xiao et~al.(2024{\natexlab{b}})Xiao, Tang, Zuo, Guo, Yang, Tang, Fu, and Han]{xiao2024duoattention}
Guangxuan Xiao, Jiaming Tang, Jingwei Zuo, Junxian Guo, Shang Yang, Haotian Tang, Yao Fu, and Song Han.
\newblock Duoattention: Efficient long-context llm inference with retrieval and streaming heads.
\newblock \emph{arXiv preprint arXiv:2410.10819}, 2024{\natexlab{b}}.

\bibitem[Xiao et~al.(2024{\natexlab{c}})Xiao, Tian, Chen, Han, and Lewis]{xiaoefficient}
Guangxuan Xiao, Yuandong Tian, Beidi Chen, Song Han, and Mike Lewis.
\newblock Efficient streaming language models with attention sinks.
\newblock In \emph{The Twelfth International Conference on Learning Representations}, 2024{\natexlab{c}}.

\bibitem[Zhang et~al.(2023)Zhang, Sheng, Zhou, Chen, Zheng, Cai, Song, Tian, R{\'e}, Barrett, et~al.]{zhang2023h2o}
Zhenyu Zhang, Ying Sheng, Tianyi Zhou, Tianlong Chen, Lianmin Zheng, Ruisi Cai, Zhao Song, Yuandong Tian, Christopher R{\'e}, Clark Barrett, et~al.
\newblock H2o: Heavy-hitter oracle for efficient generative inference of large language models.
\newblock \emph{Advances in Neural Information Processing Systems}, 36:\penalty0 34661--34710, 2023.

\bibitem[Zhou et~al.(2024)Zhou, Ning, Hong, Fu, Xu, Li, Lou, Wang, Yuan, Li, et~al.]{zhou2024survey}
Zixuan Zhou, Xuefei Ning, Ke~Hong, Tianyu Fu, Jiaming Xu, Shiyao Li, Yuming Lou, Luning Wang, Zhihang Yuan, Xiuhong Li, et~al.
\newblock A survey on efficient inference for large language models.
\newblock \emph{arXiv preprint arXiv:2404.14294}, 2024.

\end{thebibliography}
